# R&D Analyst: An Interactive Approach to Normative Decision System Model Construction

Peter J. Regan
Engineering-Economic Systems Dept.
Stanford University
Stanford, CA 94305

Samuel Holtzman
Strategic Decisions Group
2440 Sand Hill Road
Menlo Park, CA 94025

## Abstract

This paper describes the architecture of R&D Analyst, a commercial intelligent decision system for evaluating corporate research and development projects and portfolios. In analyzing projects, R&D Analyst interactively guides a user in constructing an influence diagram model for an individual research project. The system's interactive approach can be clearly explained from a blackboard system perspective. The opportunistic reasoning emphasis of blackboard systems satisfies the flexibility requirements of model construction, thereby suggesting that a similar architecture would be valuable for developing normative decision systems in other domains. Current research is aimed at extending the system architecture to explicitly consider of sequential decisions involving limited temporal, financial, and physical resources.

## 1 INTRODUCTION

This paper presents a blackboard system as an effective architecture for an intelligent decision system. Specifically, this system architecture meets the need for a mechanism to guide the user through a consultation based on decision analysis principles to construct effective decision models. This discussion is aimed at both computer scientists involved with decision systems and decision analysts.

The paper is organized as follows. Sections 2 and 3 introduce normative decision systems and blackboard systems, respectively. Section 4 presents the model-construction problem in R&D Analyst, a commercial intelligent decision system in the domain of research and development (R&D) decision-making, and describes the overall system architecture. Section 5 presents an overview of R&D Analyst, addressing the link between the organization and behavior of the system. The emphasis is on control issues related to the system architecture, rather than on details of R&D decision analysis. Sections 6 through 8 mention advanced issues, introduce some ongoing research issues, and present conclusions.

## 2 NORMATIVE DECISION SYSTEMS

"Normative decision systems" is becoming the designation for computer systems that address the tasks of modeling and solving decision problems in a particular way. Such systems are normative in that they are based on the principles of decision analysis, where probability theory is used to handle uncertainty, and maximization of expected utility is the decision criterion (Howard 1966, Howard and Matheson 1981a). Influence diagrams have been developed as a convenient and powerful normative representation language for modeling and solving decision problems (Howard and Matheson 1981b).

The basic principles of probability theory have been recognized for about two centuries (Bayes, Bernoulli, Fermat, Laplace) and those of decision analysis (DA) for several decades (Howard 1966). In contrast, research on normative decision systems began more recently (Holtzman 1989), with the first commercial systems reaching the market only in the last few years (Heckerman et al. 1990, Matheson et al. 1992 (in preparation)). Driving normative systems research is the inability of traditional expert systems to effectively handle uncertainty and the interpretational difficulties associated with nonnormative quantitative uncertainty methods (Jackson 1990, Henrion et al. 1991, Holtzman and Breese 1986).

The performance of normative decision systems has been heightened by advances in methods for probabilistic inference and evaluation of influence diagrams (Olmsted 1983, Shachter 1986, Pearl 1988, Lauritzen and Spiegelhalter 1988). Yet, while considerable effort has been devoted to the development and refinement of inference algorithms, only recently has effort been directed at constructing an appropriate influence diagram model of a real decision problem (Holtzman, 1989, Breese et al. 1991).

Intelligent decision system (IDS) technology represents one approach for computer-based model construction

(Holtzman 1989). IDS technology takes an expert systems (ES) approach to capturing the procedural knowledge of a decision analyst well versed in a given application area (domain of discourse). The decision model construction process is organized as a "coached" consultation based on DA principles.

An important distinction between an IDS and an ES is the role of expert system technology. In an ES, expert system technology is the basis for reaching a conclusion regarding the decision at hand; whereas in an IDS, expert system technology is used exclusively to construct the decision model, with inference and solution of the decision problem based firmly on strict normative theory. This normative theory is implemented as decision-analytic optimization algorithms embodied in an influence diagram evaluation system.

## 3   BLACKBOARD ARCHITECTURES

Blackboard systems have been developed to solve problems by opportunistically applying diverse types of knowledge. The goal of a blackboard system is to apply appropriate knowledge at the appropriate time. A blackboard architecture consists of three parts: knowledge sources, the blackboard data structure, and a control element.

The concept of a blackboard system is a literal implementation of the concept of a production system and was initially developed by Newell and Simon in their study of general problem-solving methods (Newell and Simon 1972). Hearsay-II, a speech-recognition system (Erman et al. 1980), was the first large blackboard system. Since its introduction, a variety of other researchers have refined and expanded the blackboard concept (Nii 1986a, 1986b; Engelmore and Morgan 1988, Jagannathan et al. 1989).

Common to the various blackboard systems is a central control cycle, which consists of the following steps:

1. Identify the eligible knowledge sources
2. Select a knowledge source for execution
3. Execute the chosen knowledge source (KS).

This cycle, identical to that in a production system, proceeds as partial solutions are refined and integrated into a complete solution. To support this cycle, KSs consist of two parts: a condition, which specifies the situations for which a KS is appropriate, and an action, which contributes to the problem-solving process by making changes to the blackboard data structure. The internal structure and representation of KSs depend on the task at hand and are not constrained by the general definition of a blackboard architecture. Typically, KS actions are procedures or collections of rules or some combination thereof.

The blackboard data structure maintains a representation, typically hierarchical, of the partial solution state, which is referred to by the KS conditions and modified by the KS actions. A control element coordinates the central control cycle and applies knowledge of its own to select a candidate KS for execution from the eligible KSs at each iteration of the cycle. The control knowledge used to determine which KS to execute must be tailored to specific domain requirements and may be responsive to the current state of the blackboard data structure.

## 4   AN OVERVIEW OF R&D ANALYST

R&D Analyst is a commercial IDS that applies DA to assist managers of corporate research and development portfolios. For a particular research project in a portfolio, the aim of the system is fourfold:

1. Construct an influence diagram (ID) decision model for the project
2. Evaluate the ID to obtain insights from sensitivities and recommendations
3. Appraise the ID for further project insights
4. Return summary information about the project to the portfolio.

The portion of the system addressed in this paper is the coached consultation process for constructing (or, more technically, formulating) a decision model tailored to a particular research project. The entire system—involving construction, evaluation, and appraisal of a portfolio of projects (individually and in aggregate)—shares the architecture described below. We focus here on knowledge-based model construction.

### 4.1   THE DECISION-MODEL-
### CONSTRUCTION PROBLEM IN R&D
### ANALYST

Constructing a decision model for a particular problem is often a complex cognitive task that involves a large number of individuals, each with particular expertise. The following are elements of a quality decision process (Creswell and McNamee 1991):

- Appropriate frame
- Creative alternatives
- Meaningful, reliable information
- Clear values and trade-offs
- Logically correct reasoning
- Commitment to action.

The challenge in constructing ID models of complex decisions is to capture the distinctions that are important to the decision-maker while adhering to DA principles. In professional consulting practice, ID model construction typically starts from scratch in a group setting moderated by a decision analyst. The decision analyst guides the decision participants through a process of eliciting expertise and making judgments. This process resembles more a meandering river than an airport runway.

The ID is constructed for a particular decision situation and is generally not used again after the decision is made. This resource-intensive process is typically justified only for important decisions involving large-scale resources. In contrast, using IDS technology we can capture DA process knowledge for a particular domain and thereby allow users to amortize their investment over many decisions, none of which might otherwise justify the cost of a full-scale review.

The feasibility of developing an IDS rests on the concept of a decision class, a group of decisions that share similar features yet are each unique in some way (Holtzman 1989). In the case of R&D Analyst, research projects share such features as technical success, R&D investment, and contribution given technical success. Notwithstanding these similarities, each R&D project is inherently unique: for example, the technical challenges of any given project are likely to be one-of-a-kind.

A critical challenge in designing an IDS is to strike a balance between maintaining flexibility and providing appropriate guidance. Flexibility is important in the forward as well as the backward direction. Forward issues include providing the user with freedom of movement in addressing various aspects of the decision and suitable modeling options for important distinctions. Backward issues include easy correction of errors and extensive revision capabilities to handle changes in problem structure and assumptions during the analysis.

## 4.2 R&D ANALYST AS A BLACKBOARD SYSTEM

The blackboard metaphor captures the essence of the interactive model construction problem. This metaphor aptly describes the manner in which a skilled decision analyst (control element and process knowledge source) guides a group of experts (domain knowledge sources) in the incremental construction of an ID representation (blackboard structure) of a decision problem. The guidance and flexibility of a decision analyst skilled in a particular class of decisions are the critical properties desired in a system architecture for IDS development.

At a conceptual level, there are important similarities between blackboard systems and traditional rule-based systems: each has a knowledge base and a control strategy for applying knowledge to solve a problem. However, we preferred the natural decomposition of R&D Analyst facilitated by a blackboard perspective into a blackboard structure consisting primarily of an augmented ID, KSs that are not rules in any traditional sense, and a highly interactive control element that guides and is influenced by the user of the system.

The examination of R&D Analyst as a blackboard system follows a format used to review other such systems (Nii 1986b).

### 4.2.1 The Task

The task under consideration is the construction of an influence diagram that captures the factors necessary for making a funding decision about an R&D project. From a purely structural perspective, the goal of the process is to construct a complete influence diagram. However, from both a decision analyst's and a user's perspective, the goal of the IDS consultation is a quality DA process. In the absence of a quality DA process that meets the decision quality criteria listed earlier, the valid influence diagram will not be persuasive for decision-making.

### 4.2.2 The Blackboard Structure

The central component of the blackboard data structure of R&D Analyst is the ID itself. For R&D Analyst, the traditional influence diagram syntax has been extended with R&D domain-specific knowledge. This additional information allows the system to respond to the current status of the modeling process with an enriched user interaction.

As the ID is constructed, data are stored to capture important information regarding the user's modeling choices. A general principle in organizing this additional information has been to maintain information as locally as possible, so that when traditional ID operations such as adding or deleting nodes are performed, the information corresponding to those nodes is also updated appropriately.

### 4.2.3 The Knowledge Sources

We have developed three types of KSs for R&D Analyst:

- Knowledge specialists
- Knowledge utilities
- Control specialists.

A key design decision was to take advantage of the hierarchical structure of the influence diagram by organizing most of the knowledge specialists according to domain-specific variable types. For instance, one KS handles the modeling of the Technical Achievement node, another handles the tasks contributing to overall technical success, and others handle the various components contributing to task success.

Eligibility and behavior of knowledge specialist KSs are context-specific. A KS for a given variable type can contribute to the model construction problem only when a node of that type is eligible for assessment in the ID (Section 4.2.4). Also, the response of a given variable type KS depends not only on the local state of a variable but also on other information available on the blackboard (Section 5.2).

Knowledge utilities are different from knowledge specialists in that they typically apply across a variety of variable types. For example, a special units management KS maintains consistency of the inputs. Units management is important to maintain the user's confidence that the system interprets quantitative information appropriately for computations. Similarly, a special KS coaches the user through a probabilistic assessment of the uncertainty associated with model variables. Finally, a set of process KSs assists the user in managing consistent revision of the model structure.

Control specialists are responsible for managing the central control cycle. They are few in number but are extremely important for efficient and effective system performance.

We focus in this paper on the behavior of KSs rather than on their detailed structure. The design and development of the underlying KS representation is examined in a related paper (Holtzman and Regan, 1992 (in preparation)).

### 4.2.4    Control

Control in R&D Analyst is guided by three interacting components:

- Influence diagram principles
- R&D decision analysis expertise
- User preferences.

These three control components are integrated to provide guidance while maintaining flexibility. The R&D Analyst model construction process is organized as an instance of the central control cycle of blackboard systems (Figure 1). The algorithm is an example of goal-driven beam search (Nilsson 1980).

The focus node stack of ID nodes is a key attention-focusing component of the central control cycle. The first element of the focus node stack is known as the focus node, and its role is to define the subdiagram currently of interest. Contributors to a focus node consist of the focus node itself and all its predecessors, whether direct or indirect.

Two examples should clarify the issue. When model construction commences with the core diagram (Figure 2), the value node is the only item in the focus node stack. The Net Present Value node is therefore the focus node, and it has five contributors (itself and four predecessors).

If the user selects the Technical Achievement node for assessment in the first iteration of the control cycle and adds two new nodes representing technical tasks, then the system pushes the Technical Achievement node onto the focus node stack, which now has two elements. Since the central control cycle considers the eligibility for assessment of only contributors (in this case the Technical Achievement node itself and the two newly added task nodes) to the focus node, the other nodes in the diagram are screened from consideration.

The focus node stack implements a mechanism recognizing that the user wants to complete related assessments and does not want to be bothered by a long list of irrelevant possibilities. The stack uses the ID structure to define relatedness of assessment goals.

Knowledge-specialist KSs are responsible for pushing ID nodes onto the focus node stack when deemed appropriate from the perspective of R&D DA expertise. The control specialist KSs are responsible for popping focus nodes from the stack when all their contributors have been completely assessed. In general, automatic focus node stack management is sufficient. At times, however, user modification of assessment precedence is useful. When all focus nodes other than the Net Present Value node have been popped from the focus node stack and no contributors to the value node are eligible for assessment, then the influence diagram model is complete.

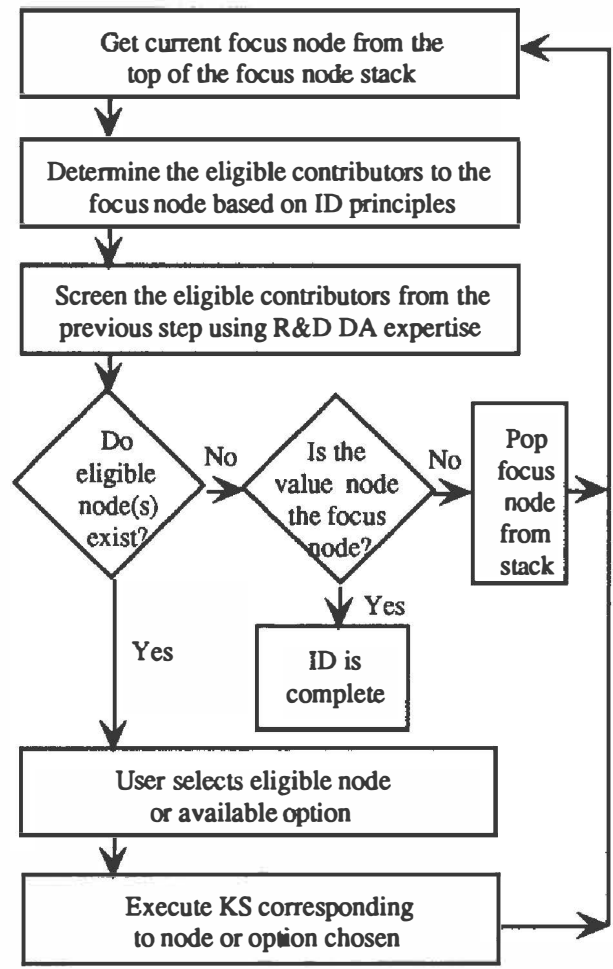

Figure 1

The system and the user share the burden of control in the R&D Analyst model construction central control cycle. Including the user in the control process is critical to the flexibility of construction. Yet the screening steps based on the focus node stack, influence diagram principles, and R&D DA expertise provide important guidance and keep the user focused.

Several user-accessible blackboard settings affect the degree of user involvement in the control process, thereby allowing users to take as much responsibility for control as they are comfortable with and are capable of handling. The system responds dynamically to these control settings as the user adjusts them throughout a consultation.

For efficiency, however, in certain situations the main control loop is bypassed when the context clearly identifies which KS to apply next. Intelligent control requires knowing when to circumvent the central control cycle (Dodhiawala 1989). These efficiency measures constitute a break from the traditional blackboard architecture in which KSs are triggered only within the central control cycle, but they effectively reduce the overhead associated with control when the benefits of control are not required.

## 5   AN OVERVIEW OF R&D ANALYST

### 5.1   THE DECISION MODEL CORE

The model construction process starts with the core diagram (Figure 2) and is complete when two criteria are met. First, the ID must be well formed in a decision analytic sense, with appropriate information provided for each node. Second, the ID must appropriately represent the user's understanding of the decision problem. The first criterion syntactic and is enforced by the structure of the consultation (e.g., the elibility criteria in the central control cycle), whereas the second is semantic and is facilitated by the content of the consultation.

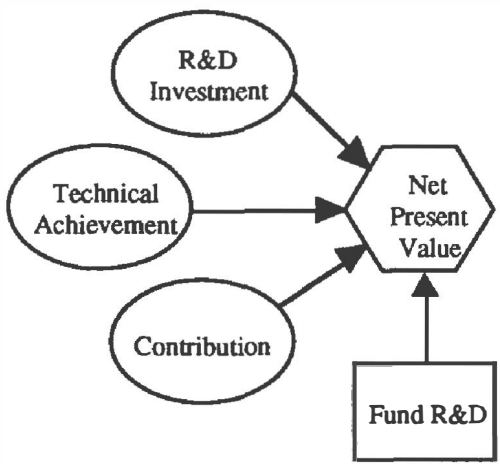

Figure 2:  Core Influence Diagram

From the perspective of model construction, an R&D Analyst consultation organizes the decision participants' knowledge about the core uncertainties (e.g., Technical Achievement, R&D Investment, Contribution).  In the rare case of a simple project with a decision-maker who is comfortable assessing these three uncertainties directly, the process is trivial.  More typically, however, the user prefers to decompose the problem into subcomponents more easily assessed by appropriate experts.

The KSs embody the DA process knowledge and R&D domain knowledge required to guide the model-construction process.  The solution process reduces the model back to the core diagram, integrating according to normative DA principles all of the information provided by the user.  Hence, the overall process corresponds to ID expansion for the convenience of introducing expertise at appropriate levels, followed by reduction for purposes of decision and insight.  Once the ID is complete, the values of the core uncertainties can be computed and returned to the portfolio of all research projects, so that various diagnostic evaluations can be performed.

The user is a key participant in the control element of the blackboard system.  The intent is to provide the feel of a traditional "transaction system" (e.g., spreadsheet) to a system that is fundamentally consultation driven (e.g., software tutorial).  In Figure 3, we see the Formulation Coach corresponding to the core diagram.

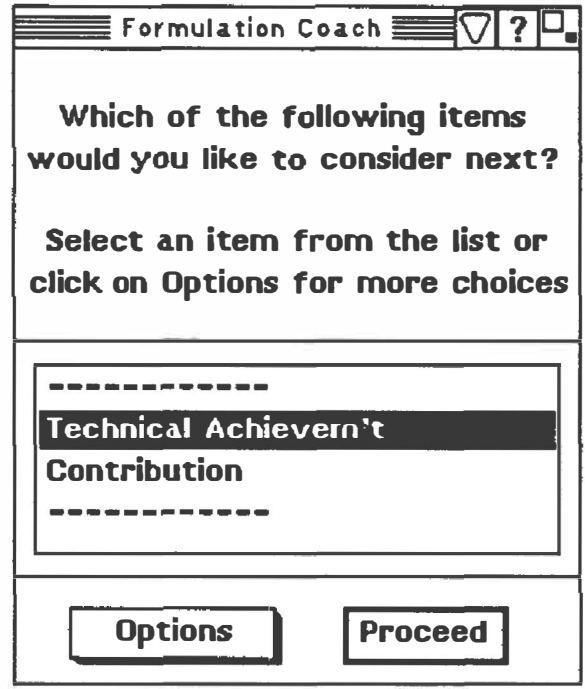

Figure 3:  Main Control Loop User Interface

This dialog box presents the user at each iteration of the control cycle with the eligible nodes for assessment along with a menu of options briefly described below.

| Save | Save the current state of the blackboard structure. |
|------|------|
| User Profile | Inspect and modify the user's level of expertise with the system. |
| Project Globals | Inspect and modify the generally available information. |
| Project Attributes | Inspect and modify the categorization of the project in the portfolio. |
| Project Units | Inspect and modify the list of units the system recognizes. |
| Redefine Project | Return to an initial analysis introduction module. |
| Evaluate TA | Calculate the probability of success implied by the current technical model. |
| Change Topic | Inspect and modify the current focus node stack. |
| Backtrack | Delete or modify part of the existing model. |
| Exit Consultation | End the consultation. |

Notice that only a subset of the nodes in the core diagram is eligible for assessment. As discussed earlier, the value node is the initial focus node, with all five nodes in the core diagram as contributors. The value node and funding decision node are screened out, since they are automatically completed by the system with no user intervention. Nor is the R&D investment node eligible. In fact, R&D investment remains screened out until the technical model has been completed, since assessing the required investment depends on the nature of the technical tasks that must be accomplished. These are simple examples of R&D DA expertise control knowledge.

If we wanted to avoid sharing control with the user, the control KSs could carry the screening process to the point where only one option were available. Such a system with limited user-based control would be more similar to traditional blackboard systems but would be inappropriate for our purposes.

## 5.2 TECHNICAL ACHIEVEMENT

Typically, the first core uncertainty to be addressed is Technical Achievement. The goal is to decompose the Technical Achievement node in a way that matches the features of the research effort and the availability of expertise. Based on R&D DA experience, the system provides the user with a few principal types of variables: tasks, success criteria, hurdles, general uncertainties, and performance variables.

Tasks have success or failure as possibilities and are used to describe the temporal and logical requirements for successful technical achievement. Tasks can be run in parallel or in sequence, and successful technical achievement may require either success in all tasks or in at least one task. Success criteria are predecessors to tasks and can be of two types: hurdles, which have success or failure possibilities, and general uncertainties, which have user-defined possibilities. Use of general uncertainties gives the user more flexibility but at the risk of creating an unmanageably large model. Performance variables are continuous measures (e.g., temperature) with a success threshold (e.g., no cracking below 450 degrees).

We will first consider the selection of Technical Achievement from the Formulation Coach agenda for assessment. Selection of technical achievement activates the KS containing the distinctions relevant to this type of variable.

The user is asked to choose between modeling the project as a single task or as multiple tasks. Explanations and examples of the distinctions presented are available to the user to make the choices meaningful. On the basis of the user's choice, the blackboard is updated, including changes to both the ID and associated data structures.

If the user chooses to introduce two technical tasks (A and B), the focus node becomes the Technical Achievement node, and the contributing nodes become the Technical Achievement node and these two tasks. The Technical Achievement node, whose distribution has been obtained in the previous step, is screened from further assessment. This is an example of ID principle control knowledge.

For assessment of task A, a different KS is activated, containing distinctions related to various options for assessing task success. The first action is to push the task A node onto the focus node stack until the assessment of task A is complete. The activated KS handles several different types of technical variables. The system selects the most appropriate behavior in a given KS based on the immediate context. This illustrates a general design principle of the system that similar distinctions be organized in a single KS.

After defining task A success in terms of two hurdles and specifying the probability of success for each hurdle, the user again is prompted to consider task A. Note that the user could have defined the hurdles indirectly in terms of yet more technical factors, but for simplicity we have selected a fairly simple model.

On the basis of the blackboard context, the KS responds differently to the selection of task A than it did the first time. Since the component hurdles of task A have been defined, the next step for task A is to specify a corresponding probability of task success given the outcomes of the hurdles. Here, we see how ID principles and R&D DA expertise are integrated to organize the process of acquiring knowledge from the decision participants.

The default choice for task A is that success of its component hurdles ensures success, with failure otherwise. Some residual uncertainty may remain, however, even if all hurdles are successful. The distribution for a case with residual uncertainty is shown in Figure 4 (Smith et al. 1992 (submitted)). Failure of task A is certain with failure in either of the predecessor hurdles. The probability of success of task A is 0.85 conditioned on success in both of the predecessor hurdles.

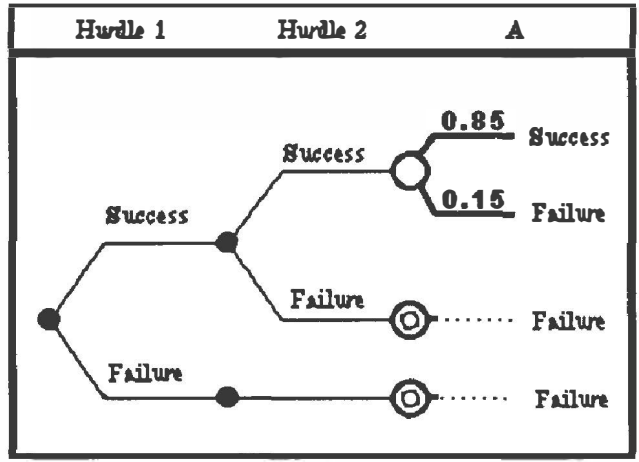

Figure 4: Conditional Distribution for a Technical Task

Following the assessment of the distribution for task A, task A is popped from the focus node stack, uncovering Technical Achievement as the focus node once again. We may then proceed to assess task B in a similar fashion, using as necessary the variables introduced for task A.

### 5.3   R&D INVESTMENT

Upon completion of the technical achievement model, the focus node is again the value node, and the user has the choice of considering the R&D investment required for the technical program or the commercial contribution given that success is achieved. This is the first time that the R&D Investment node has been eligible for assessment.

As with technical achievement, the goal of R&D investment modeling is to decompose the investment required to carry out the research. The system provides the user with a set of variable types: task investments, parameters, and general uncertainties. Task investments define the temporal cash flows associated with a given task from a library of available time series forms. Parameters are numerical arguments in task investment time series. General uncertainties have user-defined possibilities (numerical or nonnumerical) and can be used to capture relationships among the financial components of the model.

After providing investment and timing information for the two technical tasks, we are ready to consider the most complex and diverse of the core uncertainties.

### 5.4   CONTRIBUTION

In the design of KSs for model construction in the R&D domain, the strongest similarities among projects occur in the structure of the technical and investment models. For commercial contribution (given technical success), greater flexibility is essential so the user can capture the unique market features of each particular project. Thus, a larger set of variable types is available to the user, only a subset of which is presented here.

At the highest level, the system considers three components of contribution: principal product profit, related product profit, and capital investment. In practice, the nature of the "product" can differ widely, from a standard physical item to less tangible services. For this survey, we consider a simple product sales profit model, with no related products or capital investment.

Profit is modeled in terms of revenue and cost, with revenue in turn modeled as units sold and price. Here, a units manager KS handles the specification of units for sales and unit price and introduces conversion factors where necessary.

Thus far, the contribution distinctions have involved simple functional relationships. Suppose the user chooses to assess units sold directly rather than indirectly in terms of market size and share. To model a variable that changes over time, the user selects from several time series patterns supported by the system. The user then specifies whether each parameter of the time series is to be represented as certain or uncertain. This parameterization facilitates identification of key uncertainties during sensitivity analysis.

After all nodes have been fully assessed, the user is notified that the model construction process is complete. The user chooses whether to proceed to evaluate the current model or make further modifications.

## 6   ADVANCED ISSUES

A key design issue is the trade-off between guidance and flexibility. In practice, a user rarely proceeds in a straightforward linear manner from start to finish in a single session. The following are some practical concerns that motivated us to include advanced features in R&D Analyst:

- Model construction is an iterative process.
- Integration of expertise from multiple sources is critical to constructing a comprehensive model.
- The levels of guidance and flexibility should be sensitive to the skill of a particular user.
- The user may wish to consider issues in an order different from the default focus node sequence suggested by the system.
- Local information (e.g., possibilities and probabilities) may need to be updated.

- Major modeling choices (e.g., single R&D investment model versus task-by-task models) may need to be revised as new information becomes available or the problem conception changes.
- Common modeling patterns may emerge within a given company or a given sub-industry.

Most of these issues can be addressed at each iteration through the control cycle by selecting from the Options menu on the main Formulation Coach dialog box (Figure 3). Detailed discussion of these features is beyond the scope of this paper.

# 7 RESEARCH ISSUES

Several issues associated with development of normative decision systems are the subject of ongoing research.

Knowledge source organization and structure. The hierarchical structure of IDs suggests the organization of KSs according to variable types in an application domain. Our ongoing research is focused on developing general approaches for defining the appropriate grain-size of a KS and structuring the knowledge for efficient implementation and revision.

Improved knowledge-acquisition methods. Obtaining and representing expert knowledge remain bottlenecks in the normative system development process, as in other artificial intelligence systems. Fortunately, the distinctions and constructs of DA provide a structured framework in which knowledge acquisition for an IDS takes place. We are investigating improved tools for system building that involve the domain expert directly.

Resource-constrained decision-making. DA is typically applied exclusively to high-stakes business, engineering, or medical problems, primarily because of the considerable training and resources required to execute the process. An important challenge for the DA community is to develop methodologies for improving the quality of decision-making in other decision arenas.

Normative decision systems represent a medium for providing such assistance. In such systems, the cost of the system is spread over a large number of similar decisions, none of which alone would have justified decision analytic assistance. This idea has been put into practice in the development and commercialization of R&D Analyst. Another example is the Pathfinder project, which addresses pathology diagnosis (Heckerman et al. 1990), and commercial extensions of this system that address medical and engineering problems.

Building on these two examples, current thesis research is aimed at developing a normative decision system that reasons explicitly about the time and resources required to carry out various information-gathering alternatives before making a principal decision under time pressures. The research is part of an international project to develop a risk monitoring and decision system for management of offshore oil platform operations. This work builds on and extends existing methodologies as well as related research in normative decision systems (Henrion et al. 1991, Horvitz and Rutledge 1991), blackboard systems (Jagannathan et al. 1989), and artificial intelligence planning (Dean and Wellman 1991).

# 8 CONCLUSIONS

The principal conclusions for developers of normative decision systems are as follows:

- This architecture supports dynamic consultation-based influence diagram model construction without compromising the normative power of inference and solution.
- Flexibility (consultation-based) and guidance (transaction-based) must be balanced to deliver quality decision-making assistance.
- A novel aspect of R&D Analyst is the extent of user participation in the control element's conflict resolution process for selecting an appropriate knowledge source for execution.
- Extending traditional influence diagram syntax to include domain information increases dramatically their representational power.
- Consultation-based model construction systems require considerable effort at the design stage to devise an architecture and knowledge representation approach that facilitate model revision.
- Blackboard systems as incremental problem formulators are consistent with normative decision system model construction requirements.

## Acknowledgements

The authors thank Keh-Shiou Leu and Jim Matheson for their contribution to the ideas developed in this paper, and all of our colleagues at SDG who participated in the development of R&D Analyst. Peter Regan was supported during the writing of this paper by grants from the National Science Foundation (SEF-9110462) and Bureau Veritas, Paris.

## References

Breese, J.S., R. Goldman, and M.P. Wellman (1991). Knowledge-based construction of probabilistic and decision models: An overview. In *Workshop Notes from the Ninth National Conference on AI (AAAI 91): Knowledge-based construction of probabilistic and decision models*, pp. 1-17.

Creswell, D., and P. McNamee (1991). Decision quality and decision tools. *PC AI*, November/December 1991: 40-43.

Dean, T.L., and M.P. Wellman (1991). *Planning and Control*. Morgan Kaufmann, San Mateo, CA.

Dodhiawala, R.T. (1989). Blackboard systems in real-time problem solving. In Jagannathan, V., R.T. Dodhiawala, and L. Baum, eds., *Blackboard Architectures and Applications*, pp. 181-190. Academic Press, San Diego, CA.

Engelmore, R., and T. Morgan, eds. (1988). *Blackboard Systems*. Addison-Wesley, Wokingham, England.

Erman, D.L., F. Hayes-Roth, V.R. Lesser, D. Raj Reddy (1980). The HEARSAY-II speech understanding system: Integrating knowledge to resolve uncertainty. *ACM Computing Survey* 12: 213-253.

Heckerman, D., E.J. Horvitz, and B.N. Nathwani (1990). Toward normative expert systems: The Pathfinder project. Technical Report KSL-90-08, Medical Computer Science Group, Section on Medical Informatics, Stanford University, Stanford, CA.

Henrion, M., J.S. Breese, and E.J. Horvitz (1991). Decision analysis and expert systems. *AI Magazine*, Winter 1991: 64-91.

Holtzman, S. (1989). *Intelligent Decision Systems*. Addison-Wesley, Reading, MA.

Holtzman, S., and J.S. Breese (1986). Exact reasoning about uncertainty: On the design of expert systems for decision support. In Kanal, L., and J. Lemmer (eds.), *Uncertainty in Artificial Intelligence*, pp. 339-346. North Holland, Amsterdam.

Holtzman, S., and P.J. Regan (1992, in preparation). Activity graphs: A knowledge engineering tool for consultation-based systems. Strategic Decisions Group, Menlo Park, CA.

Horvitz, E.J., and D. Rutledge (1991). Time-dependent utility and action under uncertainty. In D'Ambrosio et al., eds., *Uncertainty in Artificial Intelligence: Proceedings of the Seventh Conference*, pp. 151-158. Morgan Kaufmann, San Mateo, CA.

Howard, R.A. (1966). Decision analysis: Applied decision theory. In D.B. Hertz and J. Melese, eds., *Proceedings of the Fourth International Conference on Operational Resarch*, pp. 55-77. Also in Howard, R.A., and J.E. Matheson, eds. (1981), *Readings on the Principles and Applications of Decision Analysis*, volume 1: 97-113. Strategic Decisions Group, Menlo Park, CA.

Howard, R.A., and J.E. Matheson, eds. (1981a). *Readings on the Principles and Applications of Decision Analysis*. Strategic Decisions Group, Menlo Park, CA.

Howard, R.A., and J.E. Matheson (1981b). In Howard, R.A., and J.E. Matheson, eds., *Readings on the*

*Principles and Applications of Decision Analysis*, volume 2:721-762. Strategic Decisions Group, Menlo Park, CA.

Jackson, P. (1990). *Introduction to Expert Systems*. Addison-Wesley, Wokingham, England.

Jagannathan, V., R.T. Dodhiawala, and L. Baum, eds. (1989). *Blackboard Architectures and Applications*. Academic Press, San Diego, CA.

Lauritzen, S.L., and D.J. Spiegelhalter (1988). Local computations with probabilities on graphical structures and their application to expert systems. *Journal of the Royal Statistical Society* 50: 157-224.

Matheson, J.E., S. Holtzman, and P.J. Regan (1992 in preparation). R&D Analyst: An intelligent decision system for R&D management. Strategic Decisions Group, Menlo Park, CA.

Newell, A. and H.A. Simon (1972). *Human Problem Solving*. Prentice-Hall, Englewood Cliffs, NJ.

Nii, H.P. (1986a). Blackboard systems: The blackboard model of problem solving and the evolution of blackboard architectures. *AI Magazine*, 7(2): 38-53.

Nii, H.P. (1986b). Blackboard systems: Blackboard application systems, blackboard systems from a knowledge engineering perspective. *AI Magazine*, 7(3): 82-106.

Nilsson, N.J. (1980). *Principles of Artificial Intelligence*. Tioga, Palo Alto, CA.

Olmsted, S.M. (1983). *On Representing and Solving Decision Problems*. PhD thesis, Department of Engineering-Economic Systems, Stanford University, Stanford, CA.

Pearl, J. (1988). *Probabiliistic Reasoning in Intelligent Systems: Networks of Plausible Inference*. Morgan Kaufmann, Menlo Park, CA.

Shachter, R.D. (1986). Evaluating influence diagrams. *Operations Research* 34: 871-882.

Smith, J.E., S. Holtzman, and J.E. Matheson (1992). Structuring conditional relationships in influence diagrams. Submitted to *Operations Research*. Working paper: Strategic Decisions Group, Menlo Park, CA.